\documentclass[10pt,twocolumn,letterpaper]{article}

\usepackage{cvpr}              

\usepackage{graphicx}
\usepackage{amsmath}
\usepackage{amssymb}
\usepackage{booktabs}

\usepackage{lipsum}
\usepackage{algorithm}
\usepackage{pythonhighlight}
\lstnewenvironment{PythonA}[1][]{\lstset{style=mypython, #1}}{}
\usepackage{bm}
\usepackage{multirow}
\usepackage{booktabs,tabularx}
\usepackage{siunitx}
\newcolumntype{C}{>{\centering\arraybackslash}X}

\newcommand*{\rowstyle}[1]{
  \gdef\@rowstyle{#1}%
  \@rowstyle\ignorespaces%
}

\usepackage[pagebackref,breaklinks,colorlinks]{hyperref}

\usepackage[capitalize]{cleveref}
\crefname{section}{Sec.}{Secs.}
\Crefname{section}{Section}{Sections}
\Crefname{table}{Table}{Tables}
\crefname{table}{Tab.}{Tabs.}

\def\cvprPaperID{10871} 
\def\confName{CVPR}
\def\confYear{2022}

\begin{document}

\title{MLP-3D: A MLP-like 3D Architecture with Grouped Time Mixing}

\author{Zhaofan Qiu$^{\dagger}$, Ting Yao$^{\dagger}$, Chong-Wah Ngo$^{\ddagger}$ and Tao Mei$^{\dagger}$\\
\parbox{40em}{\centering $^{\dagger}$ JD Explore Academy, Beijing, China~~~~~~~~~~~~~~$^{\ddagger}$ Singapore Management University, Singapore}\\
{\tt\small \{zhaofanqiu, tingyao.ustc\}@gmail.com, cwngo@smu.edu.sg, tmei@jd.com}
}
\maketitle

\begin{abstract}
Convolutional Neural Networks (CNNs) have been regarded as the go-to models for visual recognition. More recently, convolution-free networks, based on multi-head self-attention (MSA) or multi-layer perceptrons (MLPs), become more and more popular. Nevertheless, it is not trivial when utilizing these newly-minted networks for video recognition due to the large variations and complexities in video data. In this paper, we present MLP-3D networks, a novel MLP-like 3D architecture for video recognition. Specifically, the architecture consists of MLP-3D blocks, where each block contains one MLP applied across tokens (i.e., token-mixing MLP) and one MLP applied independently to each token (i.e., channel MLP). By deriving the novel grouped time mixing (GTM) operations, we equip the basic token-mixing MLP with the ability of temporal modeling. GTM divides the input tokens into several temporal groups and linearly maps the tokens in each group with the shared projection matrix. Furthermore, we devise several variants of GTM with different grouping strategies, and compose each variant in different blocks of MLP-3D network by greedy architecture search. Without the dependence on convolutions or attention mechanisms, our MLP-3D networks achieves 68.5\%/81.4\% top-1 accuracy on Something-Something V2 and Kinetics-400 datasets, respectively. Despite with fewer computations, the results are comparable to state-of-the-art widely-used 3D CNNs and video transformers. Source code is available at \href{https://github.com/ZhaofanQiu/MLP-3D}{https://github.com/ZhaofanQiu/MLP-3D}.
\end{abstract}

\section{Introduction}
During the past decade, the advances in Convolutional Neural Networks (CNNs) have successfully pushed the limits and improved the state-of-the-art technologies for image and video understanding \cite{he2015deep,hu2018squeeze,krizhevsky2012imagenet,simonyan2014very,qiu2017learning,carreira2017quo,tran2015learning,wang2018non,qiu2019learning,feichtenhofer2019slowfast,feichtenhofer2020x3d,qiu2021condensing,qiu2021boosting,li2021representing,qiu2017deep,yao2021seco,li2021motion}.
Besides achieving the top performances across tasks, the highly optimized implementation of convolution on various hardware makes CNNs continue to dominate computer vision research. Nevertheless, motivated by the success of attention model in Natural Language Processing (NLP) \cite{vaswani2017attention}, vision transformers \cite{dosovitskiy2020image,touvron2021training,huang2021shuffle,wang2021pyramid,liu2021swin,li2021contextual} become an alternative choice for image recognition by using multi-head self-attention (MSA) and multi-layer perceptrons (MLPs). More recently, the models solely with MLPs (i.e., MLP-like networks) without convolutions or self-attention layers are also shown able to perform well on ImageNet classification, and are more efficient for both training and inference \cite{yu2021s,chen2021cyclemlp,tolstikhin2021mlp,hou2021vision}.

\begin{figure}[!tb]
   \centering {\includegraphics[width=0.40\textwidth]{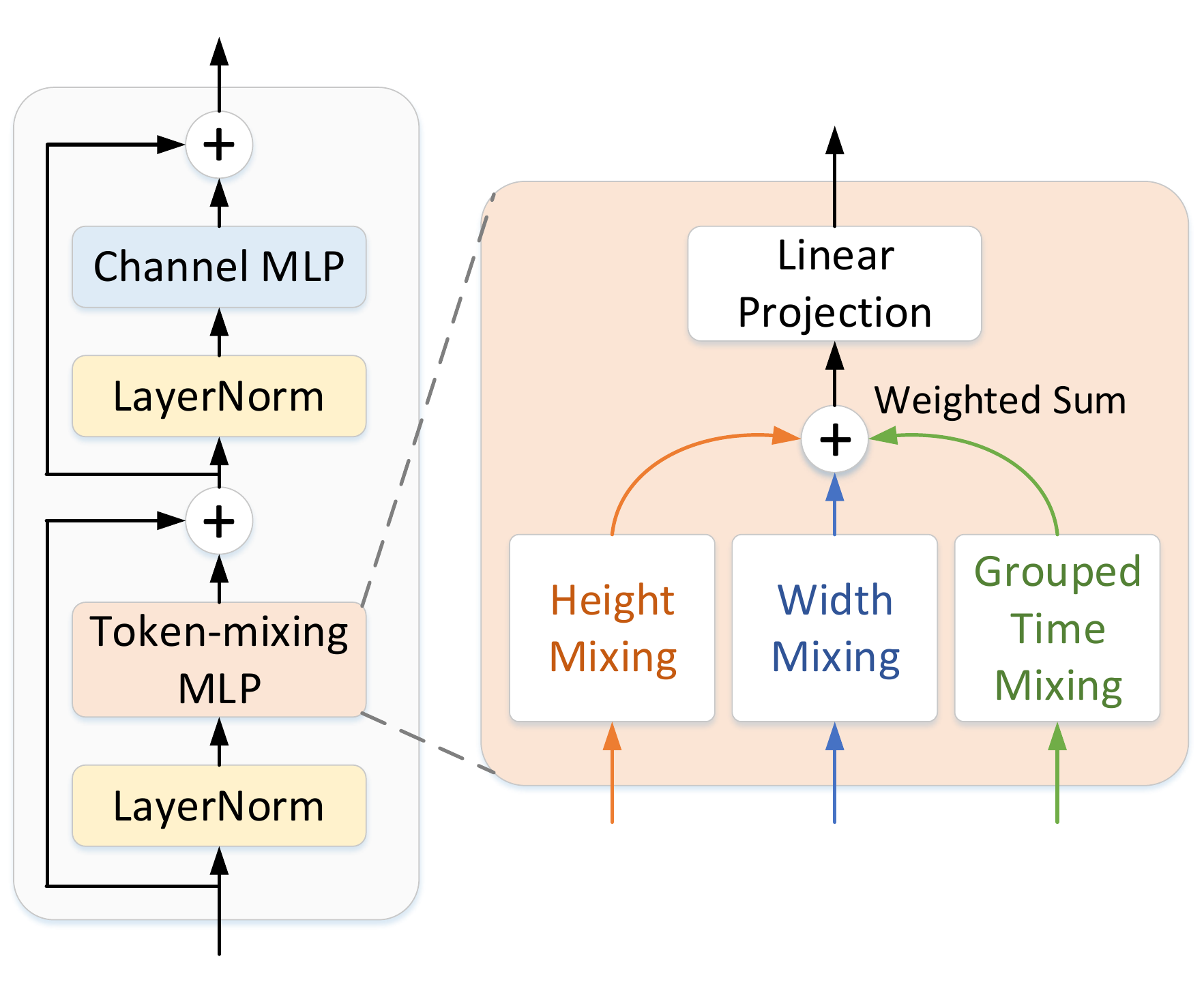}}
   \vspace{-0.1in}
   \caption{\small A schematic diagram of MLP-3D block. The block originates from the MLP-mixer layer in \cite{tolstikhin2021mlp}, and decomposes the original token-mixing MLP into three sub-operations along height, width and time dimensions, respectively. For time dimension, the novel grouped time mixing operation is exploited.}
   \label{fig:intro}
   \vspace{-0.2in}
\end{figure}

Despite having these impressive progresses on image recognition, devising MLP-like architecture on video data is seldom studied and remains challenging. The video data is more complex due to large variations in motion and rich content in visual details. Capturing useful information from such information-intensive media requires exhaustive computing resources. This property inherently poses difficulties for developing MLP-like 3D architecture from two aspects: 1) how to capture the complex temporal dynamics in videos via MLP-like operations? 2) how to reduce the expensive computations for space-time modeling?

To address the issues, in this work, we start from the design of basic MLP-style operations to model temporal sequence, and next study how to construct an efficient MLP-like architecture on the utilization of these operations. To this end, we propose MLP-3D networks - a novel MLP-like 3D architecture to model spatio-temporal dependency in videos. In MLP-3D networks, an input video clip is divided into overlapped tubelets (i.e., sequences of associated frame patches across time), and each tubelet is mapped into a visual token through a tubelet embedding layer. These tokens are then fed into several stacked MLP-3D blocks, where each block abstracts inter-token information by token-mixing MLP and intra-token information by channel MLP, as shown in \Cref{fig:intro}. The channel MLP, which shares the similar structure as the feed-forward layer in transformer \cite{vaswani2017attention}, is applied to each token independently. The token-mixing MLP is the weighted summation of three sub-operations applied across different tokens along the height, width and time dimension, respectively. The sub-operations on the spatial dimensions (height and width) follow the recipe of Cycle Fully-Connected Layer (Cycle FC) in \cite{chen2021cyclemlp}. For the time dimension, we devise a novel Grouped Time Mixing (GTM) operation, i.e., a group-based token mixing operation across tokens at different time points. By only mixing the information within each group independently, the computational complexity and the number of parameters are effectively reduced. Furthermore, based on different grouping strategies, we derive four variants of GTM operations and compose each in different MLP-3D blocks by greedy architecture search.

The main contributions of this work are summarized as follows. First, GTM is a novel family of MLP-style operations to model temporal dynamics in an economic and effective way. Second, MLP-3D networks is a new MLP-like 3D architecture by utilizing GTM operations in the decomposed token-mixing MLP. Extensive experiments conducted on Something-Something and Kinetics datasets demonstrate that MLP-3D networks achieve superior or comparable performances to widely-used 3D CNNs (e.g., SlowFast networks\cite{feichtenhofer2019slowfast}) and computationally expensive video transformers (e.g., TimeSformer \cite{bertasius2021space}). Moreover, MLP-3D networks show a great potential in developing the MLP-like architecture for video understanding.

\section{Related Work}
We group the related works into two categories: deep neural networks for image recognition and video recognition. The first category reviews the research in network design for image classification, and the second surveys a variety of video recognition models.

\textbf{Image Recognition} has received intensive research attention particularly thanks to the success of CNNs in achieving remarkable performance on several benchmarks. Significant amount of efforts are devoted to optimize CNN architectures by hand-tuning \cite{krizhevsky2012imagenet,szegedy2015going,simonyan2014very,ioffe2015batch,he2015deep,xie2017aggregated,howard2017mobilenets,zhang2018shufflenet,hu2018squeeze}.
Later, to automate the design of CNN architectures with less manual intervention, researchers have presented various Network Architecture Search (NAS) approaches, including the proposals of reinforcement learning  \cite{zoph2017neural,liu2017progressive,pham2018efficient}, architecture evolution \cite{liu2018hierarchical,luo2018neural}, and differentiable architecture search \cite{liu2018darts,chen2019progressive,li2020sgas}.

Inspired by the recent advances of attention mechanism \cite{vaswani2017attention} in NLP domain, transformer has led to a series of breakthroughs in computer vision area. The pure transformer architectures \cite{liu2021swin,wang2021pyramid,huang2021shuffle,dosovitskiy2020image} and the combinations of convolutions and transformers \cite{srinivas2021bottleneck,li2021bossnas,wu2021cvt,touvron2021going,li2021contextual} become formidable competitors to CNNs. More recently, MLP-based models \cite{yu2021s,chen2021cyclemlp,tolstikhin2021mlp,hou2021vision} are built without convolutions or attention mechanisms. Instead, MLP layers are leveraged to aggregate the spatial context over patches.

The current defacto standard for \textbf{Video Recognition} is 3D CNNs with 3D convolutions across space and time dimensions. In one of the early works \cite{ji20133d}, a 3D CNN extending directly from image-based 2D CNN is devised to recognize actions in video clip via 3D convolution. Later, there have been several attempts to improve 3D CNNs. For example, C3D \cite{tran2015learning}, by stacking several 3D convolutions and 3D poolings, demonstrates a state-of-the-art pre-training model for video understanding at the time. I3D \cite{carreira2017quo}, which pre-trains an Inception-style network \cite{szegedy2015going} on large-scale dataset, enables the extremely high fine-tuning performances on small datasets. SlowFast networks \cite{feichtenhofer2019slowfast} builds a two-path architecture consisting of a slow path with high sampling rate and a fast path with low sampling rate. In parallel with these architecture design works, the model complexity of 3D CNNs has been reduced by 3D kernel decomposition \cite{qiu2017learning,tran2018closer,xie2018rethinking} and depth-wise 3D convolutions \cite{tran2019video,feichtenhofer2020x3d}. More recently, the transformer-based architectures become a new trend of convolution-free networks on video data \cite{bertasius2021space,arnab2021vivit,liu2021video,zhang2021vidtr,neimark2021video,zhang2021token,fan2021multiscale,patrick2021keeping}.

Our work also falls into the category of convolution-free architecture for video recognition. Unlike the transformers with attention mechanisms, the token interactions in MLP-3D networks are more efficiently accomplished by fully-connected layers. Moreover, MLP-3D networks expands the research horizons of MLP-like networks to video recognition, and uniquely studies the efficient way of temporal modeling in MLP-like architecture.

\begin{figure*}[tb]
   \centering {\includegraphics[width=0.92\textwidth]{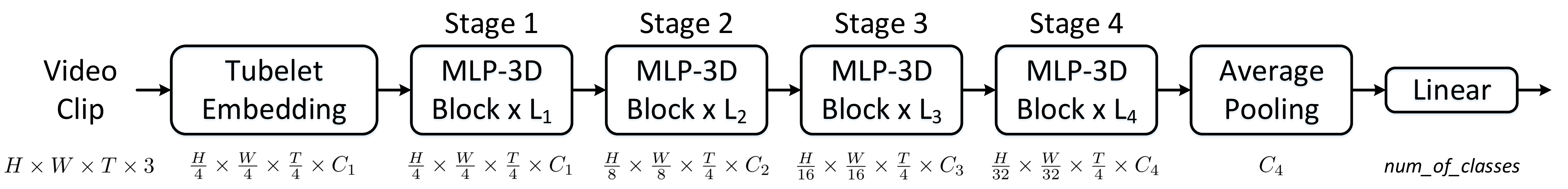}}
   \vspace{-0.1in}
   \caption{\small An overview of our proposed MLP-3D networks. $C_i$ and $L_i$ denotes the number of output channels and the repeated number of MLP-3D blocks in the $i$-th stage, respectively. The size of output feature map is also given for each block.}
   \label{fig:framework}
   \vspace{-0.2in}
\end{figure*}

\section{Our Method}
\subsection{Overall Architecture}
\Cref{fig:framework} depicts an overview of the proposed MLP-3D Networks. The basic architecture follows the philosophy of CNNs \cite{simonyan2014very,he2015deep}, where the channel dimension increases while the spatial resolution shrinks with the layer going deeper. The similar design is also exploited in hierarchical transformers \cite{liu2021swin,wang2021pyramid} and MLP-based models \cite{yu2021s,chen2021cyclemlp}.

\textbf{Tubelet embedding.} Given a video clip with the size of $H\times W\times T\times 3$, where $H$, $W$ and $T$ denotes the height, width and clip length, respectively, our models first embed the overlapped tubelets with window size $7\times 7\times 4$ and stride $4\times 4\times 4$. Each tubelet is mapped into a visual token with a higher dimension $C_{1}$ by using a shared linear embedding layer. As such, the overall tubelet embedding module produces the features with the shape of $\frac{H}{4}\times \frac{W}{4} \times \frac{T}{4}\times C_{1}$.

\textbf{Multi-stage architecture.} Then, the sequential MLP-3D blocks illustrated in \Cref{fig:framework} are applied to the tubelet tokens. The whole MLP-3D network includes four stages, and the feature resolution is maintained within each stage. A stage transition is inserted between two adjacent stages, which increases the number of channels and reduces the spatial resolution. In this way, the number of tokens from the last stage is $\frac{H}{32}\times \frac{W}{32}\times \frac{T}{4}$. The resultant tokens are finally averaged along the space and time dimensions, followed by a fully-connected layer for class prediction.

\subsection{MLP-3D Block}
The proposed MLP-3D block originates from the MLP-based block in MLP-Mixer \cite{tolstikhin2021mlp}, which replaces the multi-head self-attention module in transformer by a token-mixing MLP. In detail, MLP-based block consists of two components: channel-MLP and token-mixing MLP. \textbf{Channel-MLP} utilizes the similar structure of the feed-forward layer in transformer \cite{vaswani2017attention}, which contains two linear layers plus a GELU \cite{hendrycks2016gaussian} non-linearity in between. \textbf{Token-mixing MLP} mixes the information from tokens on different spatial/temporal positions, and characterizes the primary difference among various MLP-based models \cite{tolstikhin2021mlp,yu2021s,chen2021cyclemlp,hou2021vision}. Specifically, given the input tokens $\bm{X}$, the function of MLP-based block can be formulated as
\begin{equation}\label{eq:block}
\begin{aligned}
\bm{Y} &= \text{Token-mixing-MLP}(\text{LN}(\bm{X})) + \bm{X}, \\
\bm{Z} &= \text{Channel-MLP}(\text{LN}(\bm{Y})) + \bm{Y},
\end{aligned}
\end{equation}
where $\text{LN}$ denotes Layer Norm \cite{ba2016layer}. The output $\bm{Z}$ serves as the input to the next block until the last one.

\textbf{Decomposing token mixing.} The goal of token-mixing MLP is to capture spatial/temporal patterns by mixing the information of different tokens. Inspired by Vision Permutator \cite{hou2021vision}, MLP-3D block decomposes the token-mixing MLP and encodes the information along one axis at one time. By doing so, the token-mixing MLP can capture long-range dependencies along one dimension and meanwhile preserve precise positional information along the other dimensions. Different from \cite{hou2021vision} which decomposes the operation by height, width and channel dimensions for image recognition, MLP-3D block chooses the mixing along time axis instead of channel axis for video data. Such design shares the similar spirit as in 3D convolution decomposition \cite{qiu2017learning,xie2018rethinking,tran2018closer} and space-time divided attention \cite{bertasius2021space,arnab2021vivit}.

Concretely, the output of decomposed token mixing $\hat{\bm{Y}}$ is calculated by linearly projecting the summation of token mixing along three dimensions:
\begin{equation}\label{eq:decom}
\begin{aligned}
\hat{\bm{Y}} = \text{FC}(\bm{X}_{H}+\bm{X}_{W}+\bm{X}_{T}),
\end{aligned}
\end{equation}
where $\bm{X}_{H}$, $\bm{X}_{W}$ and $\bm{X}_{T}$ are the outputs of height, width and time mixing, respectively. $\text{FC}$ denotes a fully-connected layer. Here, we utilize the weighted summation proposed in \cite{hou2021vision} to aggregate the outputs of different mixing operations. For height/width mixing operation, we choose Cycle FC in \cite{chen2021cyclemlp}, which has been proven to be effective on capturing spatial context.

\subsection{Grouped Time Mixing (GTM)}
To further improve the efficiency of token-mixing MLP, we propose a novel Grouped Time Mixing (GTM) operation to produce $\bm{X}_{T}$ in \cref{eq:decom} by fusing the inter-token information in a grouped manner along time dimension.
Formally, we start by analyzing the simplest time mixing, which linearly maps the features of all tokens in different time points, called full time mixing. More specifically, given the reshaped input tokens as $\hat{\bm{X}}\in \mathbb{R}^{HW\times TC}$, the output of full time mixing is computed as
\begin{equation}\label{eq:time_mixing}
\begin{aligned}
\bm{X}_{T}=\hat{\bm{X}}\cdot \bm{W},
\end{aligned}
\end{equation}
where $\bm{W}\in \mathbb{R}^{TC\times TC}$ is the projection matrix. Although the operation can capture large-range dependency along the time axis, it demands geometrical progression of computational complexity $\mathcal{O}(HWT^{2}C^{2})$ and the number of parameters $\mathcal{O}(T^{2}C^{2})$ with the increase of clip length $T$.

To alleviate this limitation, we devise the Grouped Time Mixing operation, which divides the input tokens into several temporal groups and maps the tokens in each group with the shared projection parameters. As such, the computational complexity and the number of parameters are reduced because the group size is usually much smaller than the clip length. To materialize this idea, we derive four different GTM operations as depicted in \Cref{fig:gtm}, which correspond to different constructions of token groups. We detail the comparisons on the operations as follows:

\begin{figure}[!tb]
   \centering {\includegraphics[width=0.42\textwidth]{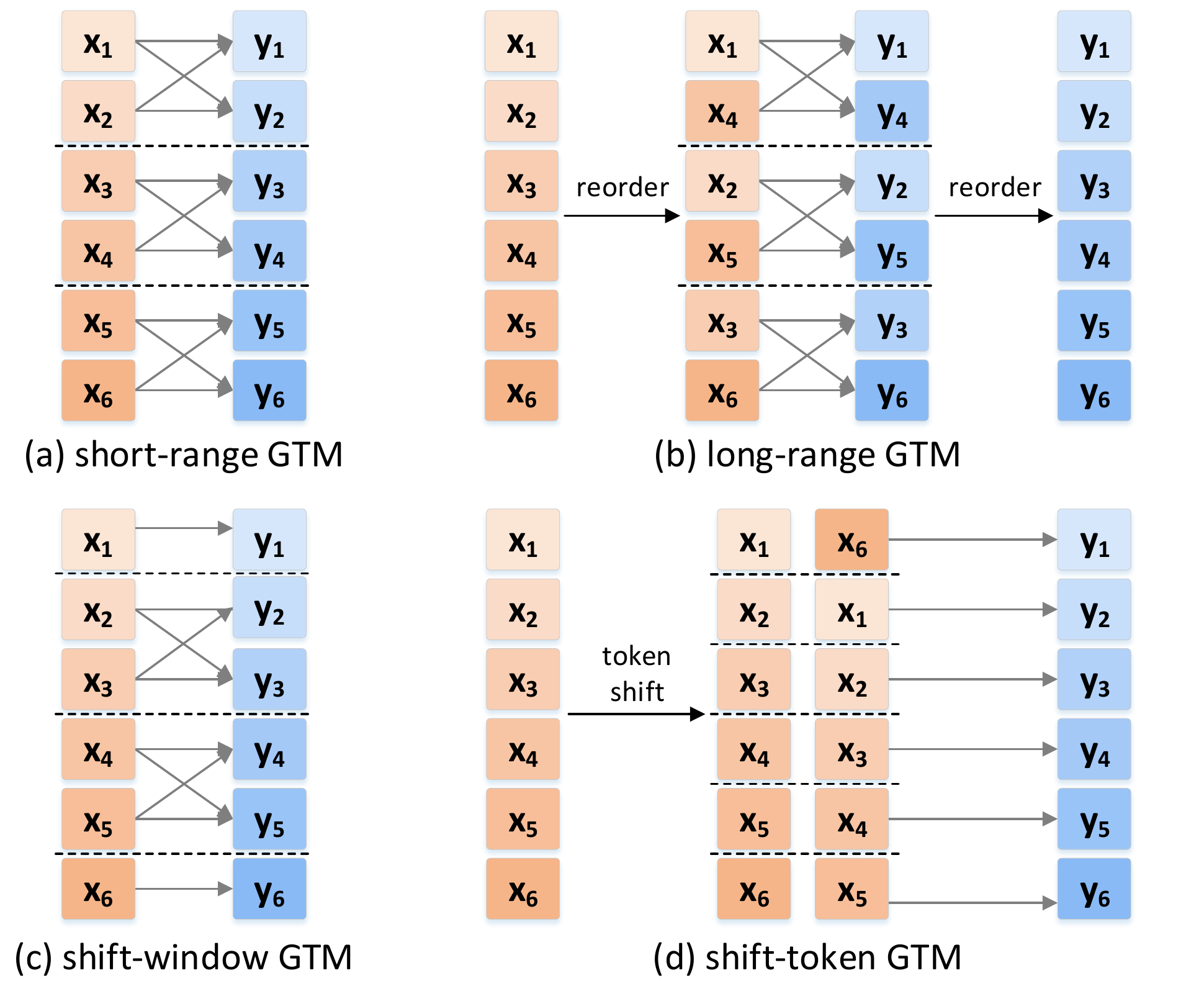}}
   \vspace{-0.1in}
   \caption{\small The illustration of four Grouped Time Mixing (GTM) operations. Each rectangle represents the input ($\bm{x}_{i}$) or output ($\bm{y}_{i}$) tokens at time point $i$. The group size is set to 2 as an example.}
   \label{fig:gtm}
   \vspace{-0.2in}
\end{figure}

\textbf{(1) Short-range GTM.} The first design evenly separates the tokens into $\frac{T}{S}$ groups, where $S$ is the group size (i.e., the number of tokens in each group). For each group, the consecutive $S$ tokens are linearly mapped by a shared matrix $\bm{W}_{S}\in \mathbb{R}^{SC\times SC}$. In other words, the short-range GTM is equivalent to making the matrix $\bm{W}$ in \cref{eq:time_mixing} be sparse:
\begin{equation}\label{eq:short_range}
\small
\begin{aligned}
\bm{W}=\begin{bmatrix}
\bm{W}_{S} & 0 & \cdots & 0 \\ 
0 & \bm{W}_{S} & \cdots & 0\\ 
\vdots & \vdots & \ddots & \vdots \\ 
0 & 0 & \cdots & \bm{W}_{S}
\end{bmatrix},
\end{aligned}
\end{equation}
in which only the values in the diagonal blocks are non-zero. As a result, the computational cost and the number of parameters is reduced to $\mathcal{O}(HWTSC^{2})$ and $\mathcal{O}(S^{2}C^{2})$, respectively. This operation is ideally similar to the window-based self-attention \cite{huang2021shuffle,liu2021swin} but remould for time mixing.

\textbf{(2) Long-range GTM.} The second design extends the first one by having an interval of $\frac{T}{S}$ time step between two consecutive tokens in each group. Such group captures long-term dependency in a video while disregarding the local patterns across adjacent frames, which is complementary to short-range GTM. The long-range GTM can be simply implemented by reordering the tokens before and after short-range GTM as shown in \Cref{fig:gtm}(b).

\textbf{(3) Shift-window GTM.} The third design is a complementary operation to short-range GTM, namely shift-window GTM. The downside of solely utilizing short-range GTM is the lack of connection across groups. Inspired by the recent successes of shifted window self-attention \cite{liu2021swin}, we shift the partition of groups in short-range GTM by an offset of $\frac{S}{2}$. Thus, alternately applying short-range GTM and shift-window GTM across different blocks in a network provides an efficient way of interaction across groups.

\textbf{(4) Shift-token GTM.}
Different from the others, the last design of shift-token GTM forms groups through shifting tokens. Specifically, with the group size $S$, each token is grouped with another $S-1$ ones, each of which reaches the reference position via shifting $1, 2,...,S-1$ time steps in a circular manner. \Cref{fig:gtm}(d) showcases an example where a token is concatenated with another token circularly shifted by one time step for linear mapping. Please note that the number of parameters of shift-token GTM is $\mathcal{O}(SC^{2})$, which is less than other GTM operations.

\textbf{Reducing parameters by weight sharing.} As per our discussion above, the number of parameters of the first three GTM operations is $\mathcal{O}(S^{2}C^{2})$ which grows fast with the increase of group size. Here, we propose to further reduce the number of parameters by sharing the projection weights between the tokens with the same time interval. Formally, the matrix $\bm{W}_{S}$ in \cref{eq:short_range} with weight sharing can be re-written~as
\begin{equation}\label{eq:relative_position}
\small
\begin{aligned}
\bm{W}_{S}=\begin{bmatrix}
\bm{w}_{0} & \bm{w}_{1} & \cdots & \bm{w}_{S-1} \\ 
\bm{w}_{-1} & \bm{w}_{0} & \cdots & \bm{w}_{S-2}\\ 
\vdots & \vdots & \ddots & \vdots \\ 
\bm{w}_{-S+1} & \bm{w}_{-S+2} & \cdots & \bm{w}_{0} \\
\end{bmatrix},
\end{aligned}
\end{equation}
where $\bm{w}_{\Delta t}\in \mathbb{R}^{C\times C}$ is the projection matrix between two tokens with $\Delta t$ time interval. The number of parameters is thus reduced to $\mathcal{O}((S\times 2-1)C^{2})$.

\begin{algorithm}[!tb]
\small
\caption{\small Codes for Grouped Time Mixing (PyTorch-like)}
\vspace{-0.05in}
\begin{PythonA}[frame=none]
# x: input tensor of shape (H, W, T, C)
# ty: mixing type, S: group size

if ty == 'shift_token':
  self.linear = nn.Linear(S*C, C)
else:
  self.linear = nn.Linear(S*C, S*C)

def grouped_time_mixing(x):
  if ty == 'short_range':
    x = self.linear(x.reshape(H, W, -1, S*C))
    x = x.reshape(H, W, T, C)
  elif ty == 'long_range':
    x = x.reshape(H, W, S, -1, C).transpose(2, 3)
    x = self.linear(x.reshape(H, W, -1, S*C))
    x = x.reshape(H, W, -1, S, C).transpose(2, 3)
    x = x.reshape(H, W, T, C)
  elif ty == 'shift_window'
    x = shift(x, S//2)
    x = self.linear(x.reshape(H, W, -1, S*C))
    x = shift(x.reshape(H, W, T, C), -S//2)
  elif ty == 'shift_token':
    x = [shift(x, i) for i in range(S)]
    x = self.linear(torch.cat(x, dim=3))
  return x
\end{PythonA}
\label{algo}
\vspace{-0.05in}
\end{algorithm}

\textbf{Implementation.} The proposed GTM operations can be readily implemented with a few lines of codes in Python. We provide an example of the codes in \Cref{algo} based on PyTorch \cite{NEURIPS2019_9015} platform. We construct the groups of tokens by calling the default \emph{reshape}, \emph{transpose}, \emph{cat} and a pre-defined \emph{shift} functions. We execute the linear mapping by the default \emph{Linear} layer.

\subsection{The MLP-3D Networks} \label{subsec:design}
In order to verify the merit of the four GTM operations, we first develop several MLP-3D network variants based on the 10-layer CycleMLP (CycleMLP-B1) \cite{chen2021cyclemlp} by replacing all the basic blocks with MLP-3D blocks which involve one certain type or two complementary types of GTM operations. Specifically, the MLP-3D network variants with a single type of GTM operations, i.e., \textbf{MLP-3D-SR}, \textbf{MLP-3D-LR} and \textbf{MLP-3D-ST}, solely utilize short-range, long-range and shift-token GTM operation, respectively. Please note that shift-window GTM is theoretically equivalent to short-range GTM when using a single type of GTM. For the variants with two mixed types of GTM operations, we exploit short-range GTM and long-range/shift-window GTM in turn for different blocks, called \textbf{MLP-3D-SR-LR}/\textbf{MLP-3D-SR-SW}, respectively. The comparisons of performance and computational cost between the basic CycleMLP-B1 and the five MLP-3D variants are discussed. Then, based on the observations from these comparisons, an oracle version of MLP-3D network is proposed for the optimal arrangement of GTM operations by a greedy search.

\textbf{Comparisons between MLP-3D network variants.} The comparisons are conducted on Something-Something V2 (SS-V2) \cite{Goyal2017TheS} dataset that is related to human-object interaction scenario and requires a precise modeling of temporal evolution. The dimension of the input video clip is set as $64\times 128\times 128$ which contains randomly cropped $128\times 128$ patches from the uniformly sampled 64 frames. For the videos with less than 64 frames, all the frames are repeated until obtaining enough frames. For each architecture, the weights are initialized with ImageNet-1K pre-trained CycleMLP-B1 model, and an extra dropout layer with 0.5 dropout rate is added before the final fully-connected layer. In the training stage, following \cite{fan2021multiscale,liu2021video}, we exploit label smoothing, random augment \cite{cubuk2020randaugment}, random erasing \cite{zhong2020random} and drop path \cite{huang2016deep} to reduce the over-fitting effect. We set each mini-batch as 512 clips, which are implemented with multiple GPUs in parallel. The network parameters are optimized by AdamW optimizer with basic learning rate of 0.0005 and weight decay of 0.05. The learning rate has one-epoch warmup and then is annealed down to zero after 32 epochs following a cosine decay.

\begin{figure}[!tb]
   \centering{\includegraphics[width=0.36\textwidth]{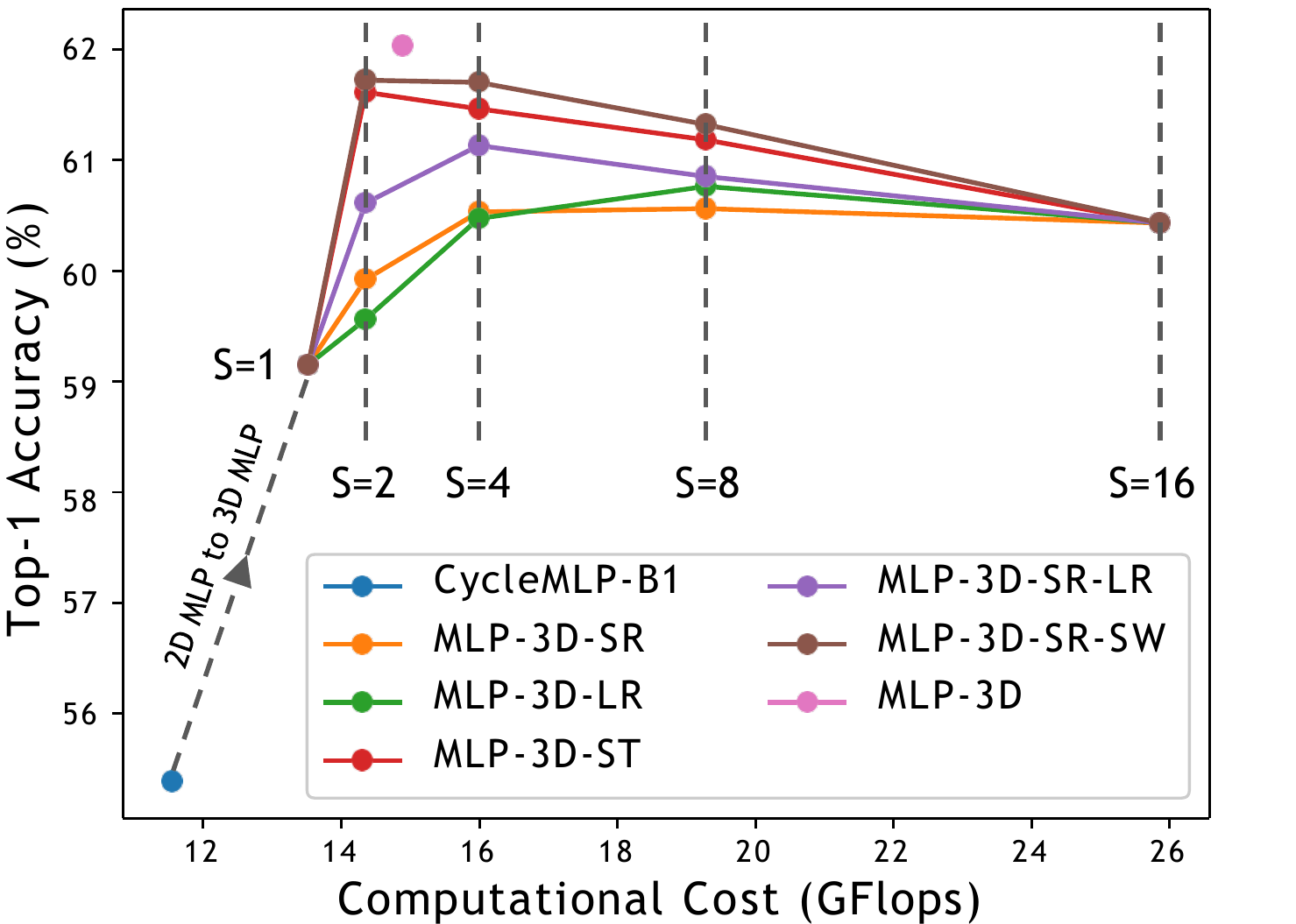}}
   \vspace{-0.1in}
   \caption{\small Comparisons of different MLP-3D network variants in terms of computational cost and top-1 accuracy on SS-V2 dataset.}
   \label{fig:com}
   \vspace{-0.2in}
\end{figure}

\Cref{fig:com} compares the performances and computations of MLP-3D network variants on SS-V2. Overall, all the MLP-3D network variants exhibit higher performance by a large margin than utilizing 2D CycleMLP-B1 on each frame independently. Specifically, among the variants with a single type of GTM operation, MLP-3D-ST with shift-token GTM achieves the best top-1 accuracy across different group size $S$. For variants with two mixed types of GTM operations, both MLP-3D-SR-LR and MLP-3D-SR-SW attain higher accuracy against solely using short-range GTM (MLP-3D-SR) or long-range GTM (MLP-3D-LR). The result basically indicates the advantage of combining different types of GTM. Moreover, for each MLP-3D network variant, the accuracy curve is like the ``$\Lambda$'' shape when $S$ varies from 1 to 16. This implies that larger group size will not always lead to higher performance, and proper sparsity constraint in GTM might benefit the network learning.

\begin{figure}[!tb]
   \centering {\includegraphics[width=0.44\textwidth]{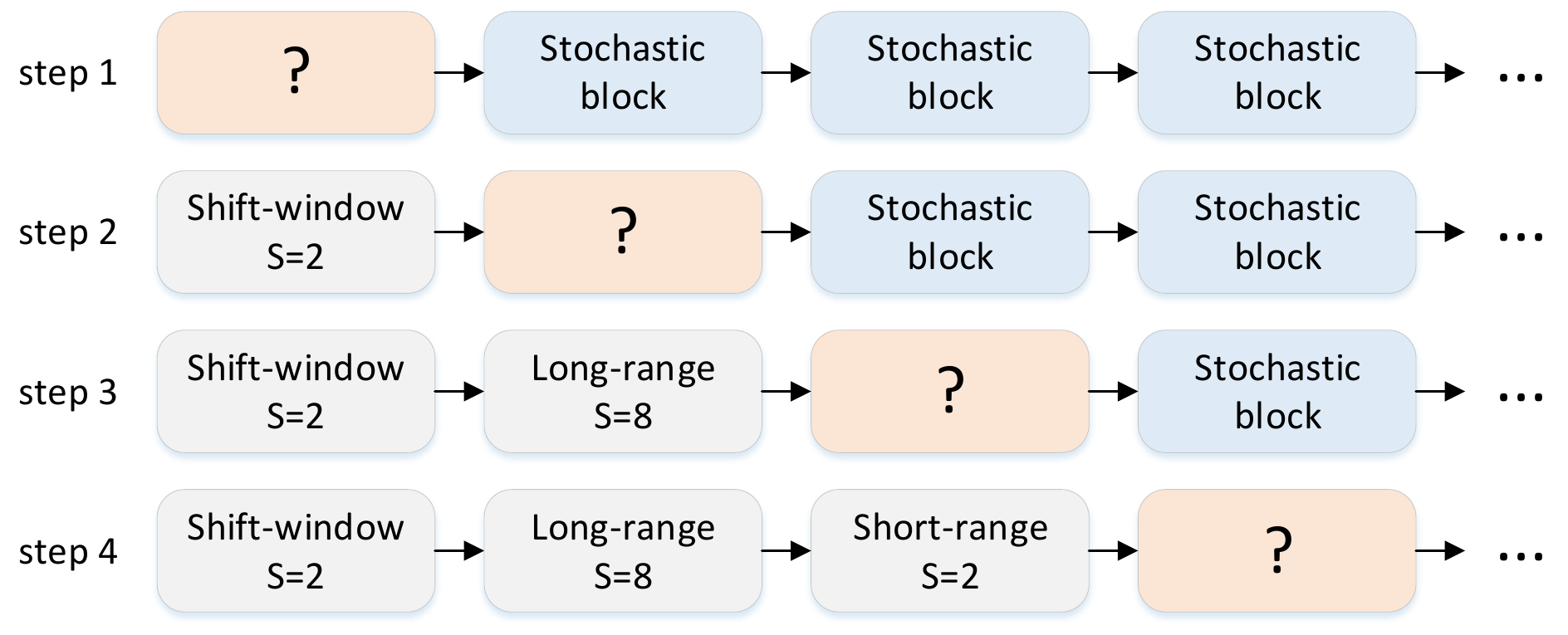}}
   \vspace{-0.1in}
   \caption{\small An example of the greedy search process to determine the MLP-3D network architecture.}
   \label{fig:search}
   \vspace{-0.2in}
\end{figure}

\textbf{MLP-3D network architecture.} Based on the empirical findings, the MLP-3D architecture can be promoted with 1) different GTM operations in different blocks; 2) carefully chosen group sizes for GTM operations; 3) proper trade-off between accuracy and computational complexity. In order to optimize these designs, we propose an efficient greedy search algorithm to determine the MLP-3D network architecture, i.e., the type of GTM operation in each block and the corresponding group size. Particularly, inspired by the high efficiency of weight sharing NAS \cite{cai2018efficient, pham2018efficient, luo2018neural}, we split the architecture search into two steps: 1) pre-training the shared weights with randomly assigned types and group sizes; 2) gradually searching the architecture with the best evaluation accuracy with regard to the pre-trained weights. For the first step, we randomly assign the types and group sizes of GTM at each iteration, and a set of time-interval-based matrices $\{\bm{w}_{\Delta t}|_{-S_{max}+1\leqslant{\Delta t}\leqslant S_{max}-1}\}$ used in \cref{eq:relative_position} is shared, where $S_{max}$ is the maximum possible group size. For the second step, the pre-trained weights are used to evaluate each architecture without additional training. In other words, given an architecture, the performance can be approximately estimated by only inferring on validation set with the shared weights. Nevertheless, it is still time-consuming to evaluate all the candidate architectures. To further reduce the time cost of architecture search, we propose to gradually determine the GTM operation of each block one-by-one. An example of the greedy search process is given in \Cref{fig:search}. At the beginning of architecture search, all the operations are randomly assigned at each forward. Then, the operation of each block is decided in turn by choosing the best-performing operation. We repeat the search process three times for more consistent results. In addition, when comparing the performances of different architectures, we further consider the computational complexity to approach a good balance. Specifically, given an architecture $\bm{\theta}$, the revised performance with the consideration of computations is given by
\begin{equation}\label{eq:objective}
\begin{aligned}
\mathcal{V}(\bm{\theta}) - \alpha \mathcal{C}(\bm{\theta}),
\end{aligned}
\end{equation}
where $\mathcal{V}(\cdot)$ and $\mathcal{C}(\cdot)$ denotes the validation accuracy and computational complexity, respectively. We set the trade-off hyper-parameters $\alpha$ as $5\mathrm{e}^{-3}$ by default. Note that the very specific MLP-3D network in \Cref{fig:com} is greedily sought by this algorithm.

\section{Experiments}
We empirically evaluate our MLP-3D networks on three challenging action recognition benchmarks:
Something-Something V1\&V2 \cite{Goyal2017TheS} and Kinetics-400 \cite{carreira2017quo}.

\subsection{Datasets}
\textbf{Something-Something} is a large-scale video dataset that focuses on human-object interaction scenario. The average video length is 4.0 seconds and all videos are captured from object-centric view with fairly clean backgrounds. The dataset contains 174 fine-grained categories of human-object interactions. The differentiation between the similar interactions is very challenging, which requires the understanding of cause-and-effect relationship in videos, e.g., ``Pushing something so that it falls off the table'' and ``Pushing something so that it almost falls off but doesn't.'' The first version (SS-V1) of the dataset contains 108K videos, which are divided into 86K, 11K, 11K for training, validation and test sets, respectively. The extended version (SS-V2) further increases the video number to 220K, which are partitioned into 170K, 25K and 25K for training, validation and test sets, respectively.

\textbf{Kinetics-400} (K-400) is a standard large-scale benchmark for video recognition, covering 400 action classes. It consists of 246K training videos, 20K validation videos and 40K test videos. Each video in the dataset is 10-second short clip trimmed from the raw YouTube video. K-400 lays particular emphasis on the visual details of objects and background instead of temporal evolution, and is usually treated as a complement to Something-Something dataset.

Note that the labels for test sets are not publicly available, and thus the performances of SS-V1, SS-V2 and Kinetics-400 are all reported on the validation set.

\subsection{Implementation Details}
\textbf{MLP-3D Networks.}
We build a family of MLP-3D networks with various model complexities, as detailed in \Cref{tab:ss}. $C_i$ and $L_i$ denote the number of output channels and the repeated number of MLP-3D block in the $i$-th stage, respectively. These settings are considered as free parameters to make the network structure tailored to the scale of video recognition problem. Here, we exploit the free parameters of CycleMLP-B1, CycleMLP-B2, CycleMLP-B3 and CycleMLP-B5 in \cite{chen2021cyclemlp}, and build a series of MLP-3D networks, namely MLP-3D-XS, MLP-3D-S, MLP-3D-M and MLP-3D-L, respectively.

\begin{table}[!tb]
\centering
\small
\caption{\small MLP-3D networks with various complexity.}
\vspace{-0.1in}
\begin{tabularx}{0.42\textwidth}{X|c|c}
\toprule
\textbf{Network} & $C_1$, $C_2$, $C_3$, $C_4$ & $L_1$, $L_2$, $L_3$, $L_4$ \\
\midrule
MLP-3D-XS & \multirow{3}{*}{64, 128, 320, 512} & 2, 2, 4, 2 \\
MLP-3D-S &  & 2, 3, 10, 3 \\
MLP-3D-M &  & 3, 4, 18, 3 \\ \midrule
MLP-3D-L & 96, 192, 384, 768 & 3, 4, 24, 3 \\
\bottomrule
\end{tabularx}
\label{tab:ss}
\vspace{-0.2in}
\end{table}

\textbf{Training stage.} The training strategies of searching and evaluating MLP-3D network variants have been described in \Cref{subsec:design}. After determining the architecture, we re-train the MLP-3D network with a similar strategy except for larger input resolution and batch augmentation \cite{hoffer2020augment}, which leads to longer training time and higher performances. In addition, considering that different GTM operations can share the same weights, we propose a novel regularization that randomly changes the type and group size of GTM to improve the generalization ability of MLP-3D networks.

\textbf{Weights initialization.} As introduced in \Cref{subsec:design}, the weights of MLP-3D networks are initialized with the ImageNet-1K pre-trained CycleMLP models. In order to maintain the semantic information of pre-trained models, center initialization in \cite{arnab2021vivit} is exploited. The idea is to copy the weights of pre-trained 2D patch embedding to the center of 3D tubelet embedding matrix. Similarly, the projection matrix $\bm{w}_{0}$ between the input and output tokens at the same time point is initialized by channel mixing operation in CycleMLP, and the other matrices $\bm{w}_{\Delta t}$ are set as zero. Such initializations make MLP-3D network perform like a 2D network when the training proceeds.

\textbf{Inference stage.} During inference, we evenly sample one/four clip(s) from each test video of Something-Something/Kinetics-400, respectively. We extract the prediction of each clip by using the three-crop strategy as in \cite{feichtenhofer2019slowfast}, which crops three square patches. The video-level prediction is obtained by averaging scores from all the clips.

\subsection{Experimental Analysis of MLP-3D Networks}

\begin{figure}[!tb]
   \centering {\includegraphics[width=0.34\textwidth]{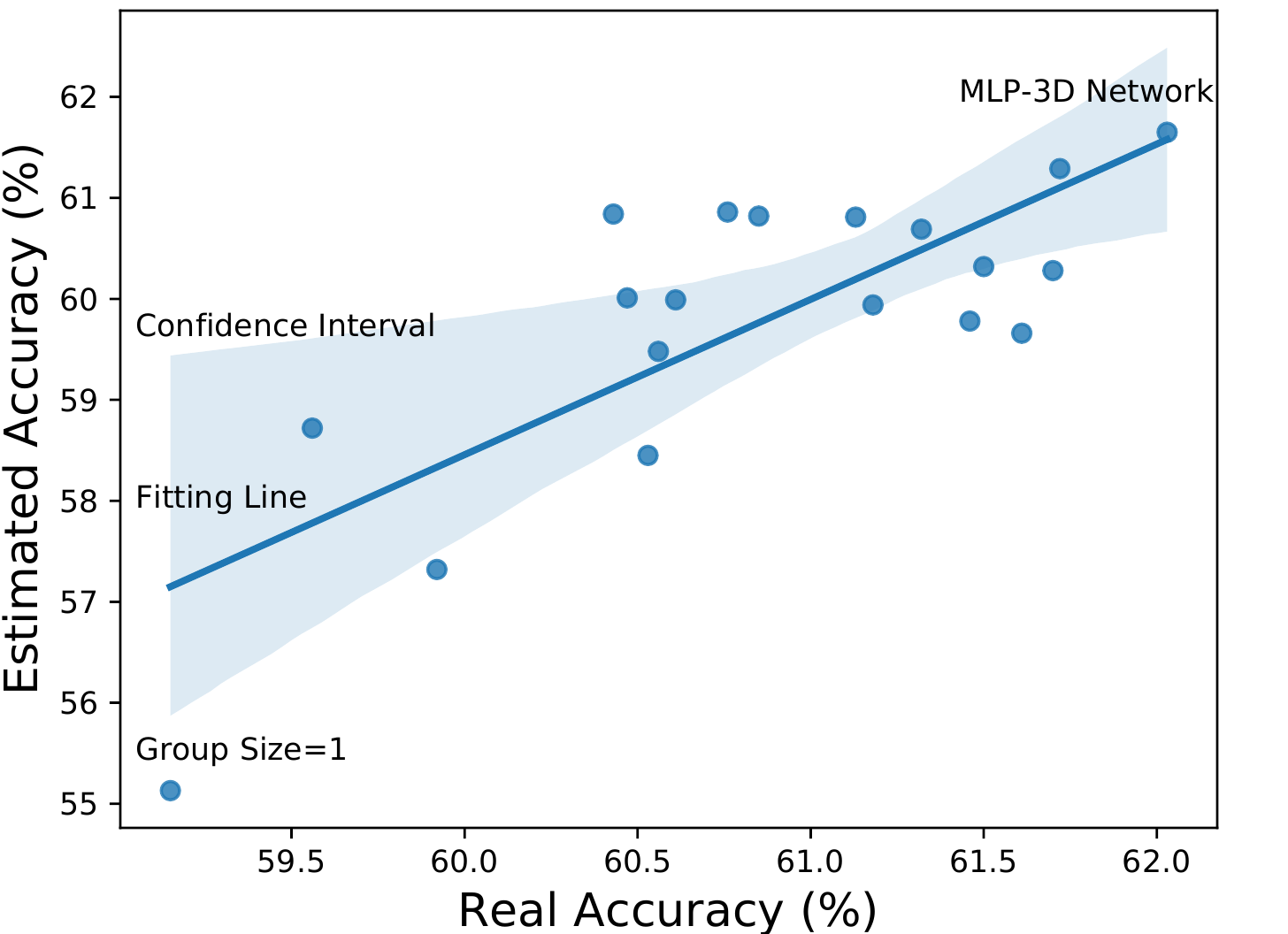}}
   \vspace{-0.1in}
   \caption{\small The visualization of real accuracy and estimated accuracy of MLP-3D-XS network variants. The fitting line of two accuracies and its confidence interval are also depicted.}
   \label{fig:relation}
   \vspace{-0.1in}
\end{figure}

\begin{figure}
\centering
   \subcaptionbox{MLP-3D-XS network on SS-V2 dataset (14.89 GFLOPs).}{
     \label{fig:network-a}
     \includegraphics[width=0.48\textwidth]{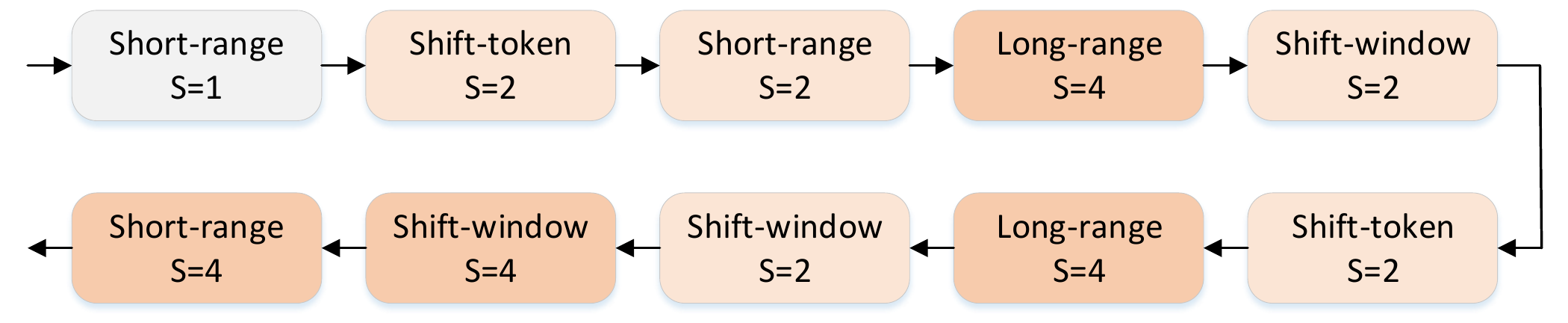}}
   \subcaptionbox{MLP-3D-XS network on Kinetics-400 dataset (14.11 GFLOPS).}{
     \label{fig:network-b}
     \includegraphics[width=0.48\textwidth]{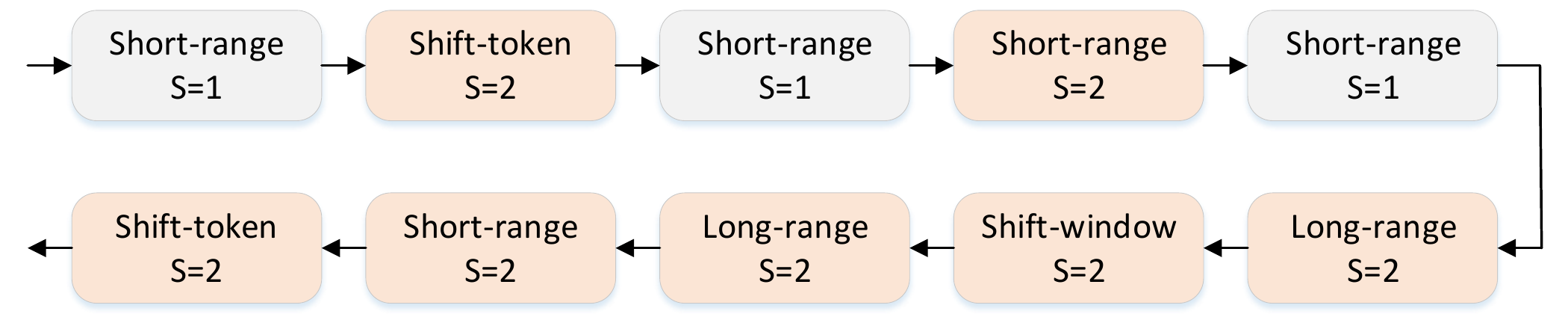}}
    \vspace{-0.1in}
   \caption{\small The MLP-3D-XS network greedily sought on (a) SS-V2 dataset, and (b) Kinetics-400 dataset, respectively. The blocks with different group sizes are in the different colors. }
   \label{fig:network}
   \vspace{-0.20in}
\end{figure}

We firstly analyze the greedy search of MLP-3D networks. \Cref{fig:relation} reveals the correlation between real accuracy and estimated accuracy using shared weights of MLP-3D-XS network variants in \Cref{subsec:design}. The real accuracy is achieved by training each architecture individually. The estimated accuracy is obtained without training by directly utilizing the pre-trained shared weights. A positive correlation is identified between these two accuracies, as shown in the figure. The result basically validates the use of estimated accuracy as an efficient approximation of real accuracy in the architecture search.

\Cref{fig:network} depicts the MLP-3D-XS networks greedily sought on SS-V2 and Kinetics-400 datasets, respectively. The optimal type of GTM and group size are given for each block. An interesting observation is that, with the same constraint of computations in \cref{eq:objective}, the network complexity searched on Kinetics-400 dataset is less than that on SS-V2 dataset. This reasonably meets our expectation since the videos in SS-V2 are known to be more complex in terms of temporal dynamics than those in Kinetics-400 dataset.

Next, we study how each design in MLP-3D networks influences the overall performance. Here, we re-train the greedily sought MLP-3D networks, i.e., from MLP-3D-XS to MLP-3D-L, with different input size ($128^2$ or $224^2$). In order to explore the effectiveness of two-path networks following \cite{feichtenhofer2019slowfast,fan2021multiscale}, we further extend the MLP-3D networks to a two-path networks by adding an additional path. To maximize the complementarity between paths, we erase the background of the input frames in the additional path by subtracting the averaged frame over time.

\begin{table}[!tb]
\centering
\scriptsize
\caption{\small Performance contribution of each design in MLP-3D networks. Top-1 accuracies are reported on SS-V2 and Kinetics-400 validation set, respectively.}
\vspace{-0.1in}
\begin{tabularx}{0.44\textwidth}{l|cc|c|C|C}
\toprule
\multirow{2}{*}{\textbf{Network}} & \multicolumn{2}{c|}{\textbf{Input size}} & \multirow{2}{*}{\textbf{Two-path}} & \multirow{2}{*}{\textbf{SS-V2}} & \multirow{2}{*}{\textbf{K-400}}\\
& $128^{2}$ & $224^{2}$ & & \\
\midrule
\multirow{4}{*}{MLP-3D-XS} & \checkmark & & & 64.6 & 75.0\\
& \checkmark & & \checkmark & 65.6 & 76.2\\
& & \checkmark & & 65.4 & 76.5\\
& & \checkmark & \checkmark & 66.0 & 77.2\\
\midrule
\multirow{4}{*}{MLP-3D-S} & \checkmark & & & 65.5 & 77.2\\
& \checkmark & & \checkmark & 66.7 & 78.0\\
& & \checkmark & & 66.7 & 79.2 \\
& & \checkmark & \checkmark & 67.2 & 80.0\\
\midrule
\multirow{4}{*}{MLP-3D-M} & \checkmark & & & 65.7 & 78.0\\
& \checkmark & & \checkmark & 66.9 & 78.8 \\
& & \checkmark & & 67.2 & 79.9\\
& & \checkmark & \checkmark & 68.0 & 81.0\\
\midrule
\multirow{4}{*}{MLP-3D-L} & \checkmark & & & 66.0 & 78.3\\
& \checkmark & & \checkmark & 66.7 & 78.9 \\
& & \checkmark & & 67.6 & 80.4 \\
& & \checkmark & \checkmark & 68.5 & 81.3\\
\bottomrule
\end{tabularx}
\label{tab:analysis}
\vspace{-0.2in}
\end{table}

\Cref{tab:analysis} details the accuracy improvements on SS-V2 and Kinetics-400 datasets by different designs in MLP-3D networks. When exploiting MLP-3D-XS as the backbone, the larger input size ($224^2$) successfully boosts up the top-1 accuracy from 64.6\% to 65.4\% on SS-V2 and from 75.0\% to 76.5\% on Kinetics-400. This demonstrates the effectiveness of training on larger resolution for video recognition. The extension to two-path networks which explores the complementarity across paths further leads to the performance improvement of 0.6\% and 0.7\% on SS-V2 and Kinetics-400, respectively. Moreover, across different MLP-3D networks, the deeper networks exhibit significantly better performance than the shallower ones. Specifically, the overall performance is improved from 66.0\% to 68.5\% on SS-V2 and from 77.2\% to 81.3\% on Kinetics-400 by replacing MLP-3D-XS with MLP-3D-L. The results verify that a deeper network has the larger learning capacity.

\subsection{Comparisons with State-of-the-Art Models}

\begin{table}[!tb]
\centering
\scriptsize
\caption{\small Comparisons with the state-of-the-art methods on SS-V2.}
\vspace{-0.1in}
\begin{tabularx}{0.48\textwidth}{l@{~}|@{~}c@{~}|@{~}c@{~}|@{~}C@{~}|@{~}c@{~}|@{~}C@{~}C}
\toprule
\textbf{Method} & \textbf{Pre-train} & \textbf{GFLOPs} & \textbf{Views} & \textbf{Params} & \textbf{Top-1} & \textbf{Top-5} \\
\midrule
TSM-RGB \cite{Lin2019TSMTS} & \multirow{8}{*}{IN-1K} & 62 & $2\times 3$ & 42.9 & 63.4 & 88.5 \\
ACTION-Net \cite{wang2021action} &  & 69 & $1\times 1$ & 28.0 & 64.0 & 89.3 \\
STM \cite{jiang2019stm} & & 66 & $10\times 3$ & 24.0 & 64.2 & 89.8 \\
SmallBig \cite{li2020smallbignet} & & 157 & $2\times 3$ & -- & 64.5 & 89.1 \\
MSNet \cite{kwon2020motionsqueeze} & & 67 & $1\times 1$ & 24.6 & 64.7 & 89.4 \\
TEA \cite{li2020tea} & & 70 & $10\times 3$ & -- & 65.1 & 89.9 \\
DG-P3D \cite{qiu2021optimization} & & 123 & $10\times 3$ & -- & 65.5 & 90.3 \\
TDN \cite{wang2021tdn} & & 132 & $1\times 1$ & -- & 66.9 & 90.9 \\
\midrule
\textcolor{gray}{TimeSformer-HR \cite{bertasius2021space}} & \multirow{2}{*}{\textcolor{gray}{IN-21K}} & \textcolor{gray}{1703} & \textcolor{gray}{$1\times3$} & \textcolor{gray}{121.4} & \textcolor{gray}{62.5} & \textcolor{gray}{--} \\
\textcolor{gray}{ViViT-L/$16\times 2$ \cite{arnab2021vivit}} & & \textcolor{gray}{903} & \textcolor{gray}{--} & \textcolor{gray}{352.1} & \textcolor{gray}{65.4} & \textcolor{gray}{89.8} \\
\midrule
\textcolor{gray}{SlowFast R101 \cite{feichtenhofer2019slowfast}} & \multirow{3}{*}{\textcolor{gray}{K-400}} & \textcolor{gray}{106} & \textcolor{gray}{$1\times3$} & \textcolor{gray}{53.3} & \textcolor{gray}{63.1} & \textcolor{gray}{87.6} \\
\textcolor{gray}{MViT-B, $64\times 3$ \cite{fan2021multiscale}} & & \textcolor{gray}{455} & \textcolor{gray}{$1\times 3$} & \textcolor{gray}{36.6} & \textcolor{gray}{67.7} & \textcolor{gray}{90.9} \\
\textcolor{gray}{Video Swin-B \cite{liu2021video}} & & \textcolor{gray}{321} & \textcolor{gray}{$1\times 3$} & \textcolor{gray}{88.8} & \textcolor{gray}{69.6} & \textcolor{gray}{92.7} \\
\midrule
\textcolor{gray}{MViT-B-24, $32\times 3$ \cite{fan2021multiscale}} & \textcolor{gray}{K-600} & \textcolor{gray}{236} & \textcolor{gray}{$1\times 3$} & \textcolor{gray}{53.2}  & \textcolor{gray}{68.7} & \textcolor{gray}{91.5} \\
\midrule
MLP-3D-XS & \multirow{4}{*}{IN-1K} & 60 & $1\times 3$ & 55.1 & 66.0 & 90.4 \\
MLP-3D-S & & 108 & $1\times 3$ & 74.1 & 67.2 & 91.3 \\
MLP-3D-M & & 183 & $1\times 3$ & 88.3 & 68.0 & 91.7 \\
MLP-3D-L & & 336 & $1\times 3$ & 149.4 & 68.5 & 92.0 \\
\bottomrule
\end{tabularx}
\label{tab:ssv2}
\vspace{-0.1in}
\end{table}

\begin{table}[!tb]
\centering
\scriptsize
\caption{\small Comparisons with the state-of-the-art methods on SS-V1.}
\vspace{-0.1in}
\begin{tabularx}{0.48\textwidth}{l@{~}|@{~}c@{~}|@{~}c@{~}|@{~}C@{~}|@{~}c@{~}|@{~}C@{~}C}
\toprule
\textbf{Method} & \textbf{Pre-train} & \textbf{GFLOPs} & \textbf{Views} & \textbf{Params} & \textbf{Top-1} & \textbf{Top-5} \\
\midrule
TSM-RGB \cite{Lin2019TSMTS} & \multirow{7}{*}{IN-1K} & 62 & $2\times 3$ & 42.9 & 47.2 & 77.1 \\
STM \cite{jiang2019stm} & & 66 & $10\times 3$ & 24.0 & 50.7 & 80.4 \\
SmallBig \cite{li2020smallbignet} & & 157 & $2\times 3$ & -- & 51.4 & 80.7 \\
MSNet \cite{kwon2020motionsqueeze} & & 67 & $1\times 1$ & 24.6 & 52.1 & 82.3 \\
TEA \cite{li2020tea} & & 70 & $10\times 3$ & -- & 52.3 & 81.9  \\
DG-P3D \cite{qiu2021optimization} & & 123 & $10\times 3$ & -- & 52.8 & 81.8 \\
TDN \cite{wang2021tdn} & & 132 & $1\times 1$ & -- & 55.3 & 83.3 \\
\midrule
MLP-3D-XS & \multirow{4}{*}{IN-1K} & 60 & $1\times 3$ & 55.1 & 54.4 & 82.5\\
MLP-3D-S & & 108 & $1\times 3$ & 74.1 & 55.2 & 83.2  \\
MLP-3D-M & & 183 & $1\times 3$ & 88.3 & 56.2 & 83.5\\
MLP-3D-L & & 336 & $1\times 3$ & 149.4 & 56.5 & 83.5\\
\bottomrule
\end{tabularx}
\label{tab:ssv1}
\vspace{-0.2in}
\end{table}

We compare with several state-of-the-art techniques on SS-V2 dataset. \Cref{tab:ssv2} summarizes the performance comparisons. The baselines are pre-trained on ImageNet-1K (IN-1K), ImageNet-21K (IN-21K), Kinetics-400 (K-400) or Kinetics-600 (K-600) datasets. The ``views'' represents the number of clips sampled from the full video during inference. Overall, the small-sized MLP-3D-S network achieves the highest top-1 accuracy of 67.2\% among the methods pre-trained on ImageNet-1K. More importantly, MLP-3D-S only spends 108G FLOPs, which is 18\% less than that of TDN. The top-1 accuracy is further improved to 68.5\% by the deeper MLP-3D-L network, which leads to the performance boost of 1.6\% over the best competitor TDN. Please note that most transformer-based models employ the pre-training on larger datasets. Nevertheless, MLP-3D-L network still outperforms TimeSformer-HR and ViViT-L by 6.0\% and 3.1\%, respectively. MViT-B and Video Swin-B, which are pre-trained with more video data, expectably obtain higher accuracy. \Cref{tab:ssv1} shows the comparisons on SS-V1 dataset. In this comparison, we merely transfer the architectures searched on SS-V2 but train the weights on SS-V1 to validate the transferability. The similar performance trends are observed on SS-V1. Specifically, MLP-3D-L attains 56.5\% top-1 accuracy, which leads the performance by 1.2\% against TDN method. The results validate the use of the searched MLP-3D networks to the dataset with similar target categories.

\begin{table}[!tb]
\centering
\scriptsize
\caption{\small Comparisons with the state-of-the-art methods on K-400.}
\vspace{-0.1in}
\begin{tabularx}{0.48\textwidth}{l@{~}|@{~}c@{~}|@{~}c@{~}|@{~}C@{~}|@{~}c@{~}|@{~}C@{~}C}
\toprule
\textbf{Method} & \textbf{Pre-train} & \textbf{GFLOPs} & \textbf{Views} & \textbf{Params} & \textbf{Top-1} & \textbf{Top-5} \\
\midrule
R(2+1)D \cite{tran2018closer} & \multirow{7}{*}{None} & 75 & $10\times 1$ & 61.8 & 72.0 & 90.0 \\
ip-CSN-152 \cite{tran2019video} &  & 109 & $10\times 3$ & 32.8 & 77.8 & 92.8 \\
CorrNet-101 \cite{wang2020video} &  & 224 & $10\times 3$ & -- & 79.8 & 93.9 \\
SlowFast R101+NL \cite{feichtenhofer2019slowfast} &  & 234 & $10\times 3$ & 59.9 & 79.8 & 93.9 \\
X3D-XXL \cite{feichtenhofer2020x3d} &  & 144 & $10\times 3$ & 20.3 & 80.4 & 94.6 \\
MViT-B, $32\times 3$ \cite{fan2021multiscale} &  & 170 & $1\times 5$ & 36.6 & 80.2 & 94.4 \\
MViT-B, $64\times 3$ \cite{fan2021multiscale} &  & 455 & $3\times 3$ & 36.6 & 81.2 & 95.1 \\ 
\midrule
I3D \cite{carreira2017quo} & \multirow{9}{*}{IN-1K} & 108 & -- & 25.0 & 72.1 & 90.3 \\
NL I3D-101 \cite{wang2018non} &  & 359 & $10\times 3$ & 61.8 & 77.7 & 93.3 \\
SmallBig \cite{li2020smallbignet} &  & 475 & $4\times 3$ & -- & 78.7 & 93.7 \\
LGD-3D \cite{qiu2019learning} &  & 195 & -- & -- & 79.4 & 94.4 \\
TDN \cite{wang2021tdn} &  & 198 & $10\times 3$ & -- & 79.4 & 94.4 \\
DG-P3D \cite{qiu2021optimization} &  & 218 & $10\times 3$ & -- & 80.5 & 94.6 \\
Video Swin-T\cite{liu2021video} &  & 88 & $4\times 3$ & 28.2 & 78.8 & 93.6 \\ 
Video Swin-S\cite{liu2021video} &  & 166 & $4\times 3$ & 49.8 & 80.6 & 94.5 \\
Video Swin-B\cite{liu2021video} &  & 282 & $4\times 3$ & 88.1 & 80.6 & 94.6 \\ 
\midrule
\textcolor{gray}{ViT-VTN \cite{neimark2021video}} & \multirow{7}{*}{\textcolor{gray}{IN-21K}} & \textcolor{gray}{4218} & \textcolor{gray}{$1\times 1$} & \textcolor{gray}{11.0} & \textcolor{gray}{78.6} & \textcolor{gray}{93.7} \\
\textcolor{gray}{TokShift \cite{zhang2021token}} &  & \textcolor{gray}{2096} & \textcolor{gray}{$10\times 3$} & \textcolor{gray}{303.4} & \textcolor{gray}{80.4} & \textcolor{gray}{94.4} \\
\textcolor{gray}{TimeSformer-L \cite{bertasius2021space}} &  & \textcolor{gray}{2380} & \textcolor{gray}{$1\times3$} & \textcolor{gray}{121.4} & \textcolor{gray}{80.7} & \textcolor{gray}{94.7} \\
\textcolor{gray}{ViViT-L/$16\times 2$ \cite{arnab2021vivit}} &  & \textcolor{gray}{1446} & \textcolor{gray}{$4\times 3$} & \textcolor{gray}{310.8} & \textcolor{gray}{80.6} & \textcolor{gray}{94.7} \\
\textcolor{gray}{ViViT-L/$16\times 2$ 320\cite{arnab2021vivit}} &  & \textcolor{gray}{3992} & \textcolor{gray}{$4\times 3$} & \textcolor{gray}{310.8} & \textcolor{gray}{81.3} & \textcolor{gray}{94.7} \\
\textcolor{gray}{Video Swin-B\cite{liu2021video}} &  & \textcolor{gray}{282} & \textcolor{gray}{$4\times 3$} & \textcolor{gray}{88.1} & \textcolor{gray}{82.7} & \textcolor{gray}{95.5} \\ 
\textcolor{gray}{Video Swin-L(384$\uparrow$)\cite{liu2021video}} &  & \textcolor{gray}{2107} & \textcolor{gray}{$10\times 5$} & \textcolor{gray}{200.0} & \textcolor{gray}{84.9} & \textcolor{gray}{96.7} \\
\midrule
MLP-3D-XS & \multirow{4}{*}{IN-1K} & 57 & $4\times 3$ & 50.1 & 77.2 & 93.1  \\
MLP-3D-S &  & 102 & $4\times 3$ & 68.5 & 80.2 & 93.8 \\
MLP-3D-M &  & 170 & $4\times 3$ & 80.5 & 81.0 & 94.9 \\
MLP-3D-L &  & 308 & $4\times 3$ & 135.6 & 81.4 & 95.2 \\
\bottomrule
\end{tabularx}
\label{tab:k400}
\vspace{-0.2in}
\end{table}

Then, we turn to evaluate MLP-3D networks on the large-scale Kinetics-400 dataset. The performance comparisons are reported in \Cref{tab:k400}. Specifically, with ImageNet-1K pre-training, MLP-3D-L network achieves 81.4\% top-1 accuracy, making the improvements over the recent approaches Video Swin-B, DG-P3D, TDN and LGD-3D by 0.8\%, 0.9\%, 2.0\% and 2.0\%, respectively. Furthermore, MLP-3D-L with less FLOPs is impressively superior to several video transformers pre-trained on ImageNet-21K, e.g., ViT-VTN, TokShift, TimeSformer-L and ViViT-L, which spend about ten times more FLOPs.

\section{Conclusion and Discussion}
We have proposed a new family of MLP-like 3D architectures named MLP-3D networks for video recognition.
Particularly, we investigate the token interaction across time in MLP-like architecture, by designing MLP-3D blocks with token-mixing MLP decomposed by height, width and time dimensions. For time dimension, we have devised variants of novel grouped time mixing (GTM) operations for group-based interaction between tokens. The type of GTM and its group size of each block are determined by an efficient greedy architecture search. Experiments conducted on three datasets, i.e., Something-Something V1 \& V2, and Kinetics-400, validate that MLP-3D networks achieve superior performances than other video recognition techniques under the same pre-training scheme. The competitive performances also show the high potential of MLP-like architectures for video analysis. More remarkably, the network is easier to train and consumes less FLOPs.

\textbf{Broader Impact.} Our MLP-3D shows a great potential of MLP-like architectures for video analysis, which are easily developed with less computations. This could increase the risk of video understanding model or its outputs being used incorrectly, such as for unauthorized surveillance.

\textbf{Acknowledgments.} This work was supported by the National Key R\&D Program of China under Grant No. 2020AAA0108600. 

{\small
\bibliographystyle{ieee_fullname}
\bibliography{egbib}

\begin{thebibliography}{10}\itemsep=-1pt

\bibitem{arnab2021vivit}
Anurag Arnab, Mostafa Dehghani, Georg Heigold, Chen Sun, Mario Lu{\v{c}}i{\'c},
  and Cordelia Schmid.
\newblock Vivit: A video vision transformer.
\newblock {\em arXiv preprint arXiv:2103.15691}, 2021.

\bibitem{ba2016layer}
Jimmy~Lei Ba, Jamie~Ryan Kiros, and Geoffrey~E Hinton.
\newblock Layer normalization.
\newblock {\em arXiv preprint arXiv:1607.06450}, 2016.

\bibitem{bertasius2021space}
Gedas Bertasius, Heng Wang, and Lorenzo Torresani.
\newblock Is space-time attention all you need for video understanding?
\newblock In {\em ICML}, 2021.

\bibitem{cai2018efficient}
Han Cai, Tianyao Chen, Weinan Zhang, Yong Yu, and Jun Wang.
\newblock Efficient architecture search by network transformation.
\newblock In {\em AAAI}, 2018.

\bibitem{carreira2017quo}
Jo{\~a}o Carreira and Andrew Zisserman.
\newblock Quo vadis, action recognition? a new model and the kinetics dataset.
\newblock In {\em CVPR}, 2017.

\bibitem{chen2021cyclemlp}
Shoufa Chen, Enze Xie, Chongjian Ge, Ding Liang, and Ping Luo.
\newblock Cyclemlp: A mlp-like architecture for dense prediction.
\newblock {\em arXiv preprint arXiv:2107.10224}, 2021.

\bibitem{chen2019progressive}
Xin Chen, Lingxi Xie, Jun Wu, and Qi Tian.
\newblock Progressive differentiable architecture search: Bridging the depth
  gap between search and evaluation.
\newblock In {\em ICCV}, 2019.

\bibitem{cubuk2020randaugment}
Ekin~D Cubuk, Barret Zoph, Jonathon Shlens, and Quoc~V Le.
\newblock Randaugment: Practical automated data augmentation with a reduced
  search space.
\newblock In {\em NeurIPS}, 2020.

\bibitem{dosovitskiy2020image}
Alexey Dosovitskiy, Lucas Beyer, Alexander Kolesnikov, Dirk Weissenborn,
  Xiaohua Zhai, Thomas Unterthiner, Mostafa Dehghani, Matthias Minderer, Georg
  Heigold, Sylvain Gelly, et~al.
\newblock An image is worth 16x16 words: Transformers for image recognition at
  scale.
\newblock In {\em ICLR}, 2021.

\bibitem{fan2021multiscale}
Haoqi Fan, Bo Xiong, Karttikeya Mangalam, Yanghao Li, Zhicheng Yan, Jitendra
  Malik, and Christoph Feichtenhofer.
\newblock Multiscale vision transformers.
\newblock {\em arXiv preprint arXiv:2104.11227}, 2021.

\bibitem{feichtenhofer2020x3d}
Christoph Feichtenhofer.
\newblock X3d: Expanding architectures for efficient video recognition.
\newblock In {\em CVPR}, 2020.

\bibitem{feichtenhofer2019slowfast}
Christoph Feichtenhofer, Haoqi Fan, Jitendra Malik, and Kaiming He.
\newblock Slowfast networks for video recognition.
\newblock In {\em ICCV}, 2019.

\bibitem{Goyal2017TheS}
Raghav Goyal, Samira~Ebrahimi Kahou, Vincent Michalski, Joanna Materzynska,
  Susanne Westphal, Heuna Kim, Valentin Haenel, Ingo Fr{\"u}nd, and \etal.
\newblock The “something something” video database for learning and
  evaluating visual common sense.
\newblock In {\em ICCV}, 2017.

\bibitem{he2015deep}
Kaiming He, Xiangyu Zhang, Shaoqing Ren, and Jian Sun.
\newblock Deep residual learning for image recognition.
\newblock In {\em CVPR}, 2016.

\bibitem{hendrycks2016gaussian}
Dan Hendrycks and Kevin Gimpel.
\newblock Gaussian error linear units (gelus).
\newblock {\em arXiv preprint arXiv:1606.08415}, 2016.

\bibitem{hoffer2020augment}
Elad Hoffer, Tal Ben-Nun, Itay Hubara, Niv Giladi, Torsten Hoefler, and Daniel
  Soudry.
\newblock Augment your batch: Improving generalization through instance
  repetition.
\newblock In {\em CVPR}, 2020.

\bibitem{hou2021vision}
Qibin Hou, Zihang Jiang, Li Yuan, Ming-Ming Cheng, Shuicheng Yan, and Jiashi
  Feng.
\newblock Vision permutator: A permutable mlp-like architecture for visual
  recognition.
\newblock {\em arXiv preprint arXiv:2106.12368}, 2021.

\bibitem{howard2017mobilenets}
Andrew~G Howard, Menglong Zhu, Bo Chen, Dmitry Kalenichenko, Weijun Wang,
  Tobias Weyand, Marco Andreetto, and Hartwig Adam.
\newblock Mobilenets: Efficient convolutional neural networks for mobile vision
  applications.
\newblock {\em arXiv preprint arXiv:1704.04861}, 2017.

\bibitem{hu2018squeeze}
Jie Hu, Li Shen, and Gang Sun.
\newblock Squeeze-and-excitation networks.
\newblock In {\em CVPR}, 2018.

\bibitem{huang2016deep}
Gao Huang, Yu Sun, Zhuang Liu, Daniel Sedra, and Kilian~Q Weinberger.
\newblock Deep networks with stochastic depth.
\newblock In {\em ECCV}, 2016.

\bibitem{huang2021shuffle}
Zilong Huang, Youcheng Ben, Guozhong Luo, Pei Cheng, Gang Yu, and Bin Fu.
\newblock Shuffle transformer: Rethinking spatial shuffle for vision
  transformer.
\newblock {\em arXiv preprint arXiv:2106.03650}, 2021.

\bibitem{ioffe2015batch}
Sergey Ioffe and Christian Szegedy.
\newblock Batch normalization: Accelerating deep network training by reducing
  internal covariate shift.
\newblock {\em arXiv preprint arXiv:1502.03167}, 2015.

\bibitem{ji20133d}
Shuiwang Ji, Wei Xu, Ming Yang, and Kai Yu.
\newblock 3d convolutional neural networks for human action recognition.
\newblock {\em IEEE Trans. on PAMI}, 35(1):221--231, 2013.

\bibitem{jiang2019stm}
Boyuan Jiang, MengMeng Wang, Weihao Gan, Wei Wu, and Junjie Yan.
\newblock Stm: Spatiotemporal and motion encoding for action recognition.
\newblock In {\em ICCV}, 2019.

\bibitem{krizhevsky2012imagenet}
Alex Krizhevsky, Ilya Sutskever, and Geoffrey~E Hinton.
\newblock Imagenet classification with deep convolutional neural networks.
\newblock In {\em NIPS}, 2012.

\bibitem{kwon2020motionsqueeze}
Heeseung Kwon, Manjin Kim, Suha Kwak, and Minsu Cho.
\newblock Motionsqueeze: Neural motion feature learning for video
  understanding.
\newblock In {\em ECCV}, 2020.

\bibitem{li2021bossnas}
Changlin Li, Tao Tang, Guangrun Wang, Jiefeng Peng, Bing Wang, Xiaodan Liang,
  and Xiaojun Chang.
\newblock Bossnas: Exploring hybrid cnn-transformers with block-wisely
  self-supervised neural architecture search.
\newblock {\em arXiv preprint arXiv:2103.12424}, 2021.

\bibitem{li2021representing}
Dong Li, Zhaofan Qiu, Yingwei Pan, Ting Yao, Houqiang Li, and Tao Mei.
\newblock Representing videos as discriminative sub-graphs for action
  recognition.
\newblock In {\em CVPR}, 2021.

\bibitem{li2020sgas}
Guohao Li, Guocheng Qian, Itzel~C Delgadillo, Matthias Muller, Ali Thabet, and
  Bernard Ghanem.
\newblock Sgas: Sequential greedy architecture search.
\newblock In {\em CVPR}, 2020.

\bibitem{li2021motion}
Rui Li, Yiheng Zhang, Zhaofan Qiu, Ting Yao, Dong Liu, and Tao Mei.
\newblock Motion-focused contrastive learning of video representations.
\newblock In {\em ICCV}, 2021.

\bibitem{li2020smallbignet}
Xianhang Li, Yali Wang, Zhipeng Zhou, and Yu Qiao.
\newblock Smallbignet: Integrating core and contextual views for video
  classification.
\newblock In {\em CVPR}, 2020.

\bibitem{li2020tea}
Yan Li, Bin Ji, Xintian Shi, Jianguo Zhang, Bin Kang, and Limin Wang.
\newblock Tea: Temporal excitation and aggregation for action recognition.
\newblock In {\em CVPR}, 2020.

\bibitem{li2021contextual}
Yehao Li, Ting Yao, Yingwei Pan, and Tao Mei.
\newblock Contextual transformer networks for visual recognition.
\newblock {\em IEEE Trans. on PAMI}, 2022.

\bibitem{Lin2019TSMTS}
Ji Lin, Chuang Gan, and Song Han.
\newblock Tsm: Temporal shift module for efficient video understanding.
\newblock In {\em ICCV}, 2019.

\bibitem{liu2017progressive}
Chenxi Liu, Barret Zoph, Jonathon Shlens, Wei Hua, Li-Jia Li, Li Fei-Fei, Alan
  Yuille, Jonathan Huang, and Kevin Murphy.
\newblock Progressive neural architecture search.
\newblock In {\em ECCV}, 2018.

\bibitem{liu2018hierarchical}
Hanxiao Liu, Karen Simonyan, Oriol Vinyals, Chrisantha Fernando, and Koray
  Kavukcuoglu.
\newblock Hierarchical representations for efficient architecture search.
\newblock In {\em ICLR}, 2018.

\bibitem{liu2018darts}
Hanxiao Liu, Karen Simonyan, and Yiming Yang.
\newblock Darts: Differentiable architecture search.
\newblock In {\em ICLR}, 2019.

\bibitem{liu2021swin}
Ze Liu, Yutong Lin, Yue Cao, Han Hu, Yixuan Wei, Zheng Zhang, Stephen Lin, and
  Baining Guo.
\newblock Swin transformer: Hierarchical vision transformer using shifted
  windows.
\newblock In {\em ICCV}, 2021.

\bibitem{liu2021video}
Ze Liu, Jia Ning, Yue Cao, Yixuan Wei, Zheng Zhang, Stephen Lin, and Han Hu.
\newblock Video swin transformer.
\newblock {\em arXiv preprint arXiv:2106.13230}, 2021.

\bibitem{luo2018neural}
Renqian Luo, Fei Tian, Tao Qin, Enhong Chen, and Tie-Yan Liu.
\newblock Neural architecture optimization.
\newblock In {\em NIPS}, 2018.

\bibitem{neimark2021video}
Daniel Neimark, Omri Bar, Maya Zohar, and Dotan Asselmann.
\newblock Video transformer network.
\newblock {\em arXiv preprint arXiv:2102.00719}, 2021.

\bibitem{NEURIPS2019_9015}
Adam Paszke, Sam Gross, Francisco Massa, Adam Lerer, James Bradbury, Gregory
  Chanan, Trevor Killeen, Zeming Lin, Natalia Gimelshein, Luca Antiga, Alban
  Desmaison, Andreas Kopf, Edward Yang, Zachary DeVito, Martin Raison, et~al.
\newblock Pytorch: An imperative style, high-performance deep learning library.
\newblock In {\em NeurIPS}, 2019.

\bibitem{patrick2021keeping}
Mandela Patrick, Dylan Campbell, Yuki~M Asano, Ishan Misra~Florian Metze,
  Christoph Feichtenhofer, Andrea Vedaldi, Jo Henriques, et~al.
\newblock Keeping your eye on the ball: Trajectory attention in video
  transformers.
\newblock In {\em NeurIPS}, 2021.

\bibitem{pham2018efficient}
Hieu Pham, Melody~Y Guan, Barret Zoph, Quoc~V Le, and Jeff Dean.
\newblock Efficient neural architecture search via parameter sharing.
\newblock In {\em ICML}, 2018.

\bibitem{qiu2017deep}
Zhaofan Qiu, Ting Yao, and Tao Mei.
\newblock Deep quantization: Encoding convolutional activations with deep
  generative model.
\newblock In {\em CVPR}, 2017.

\bibitem{qiu2017learning}
Zhaofan Qiu, Ting Yao, and Tao Mei.
\newblock Learning spatio-temporal representation with pseudo-3d residual
  networks.
\newblock In {\em ICCV}, 2017.

\bibitem{qiu2021optimization}
Zhaofan Qiu, Ting Yao, Chong-Wah Ngo, and Tao Mei.
\newblock Optimization planning for 3d convnets.
\newblock In {\em ICML}, 2021.

\bibitem{qiu2019learning}
Zhaofan Qiu, Ting Yao, Chong-Wah Ngo, Xinmei Tian, and Tao Mei.
\newblock Learning spatio-temporal representation with local and global
  diffusion.
\newblock In {\em CVPR}, 2019.

\bibitem{qiu2021boosting}
Zhaofan Qiu, Ting Yao, Chong-Wah Ngo, Xiao-Ping Zhang, Dong Wu, and Tao Mei.
\newblock Boosting video representation learning with multi-faceted
  integration.
\newblock In {\em CVPR}, 2021.

\bibitem{qiu2021condensing}
Zhaofan Qiu, Ting Yao, Yan Shu, Chong-Wah Ngo, and Tao Mei.
\newblock Condensing a sequence to one informative frame for video recognition.
\newblock In {\em ICCV}, 2021.

\bibitem{simonyan2014very}
Karen Simonyan and Andrew Zisserman.
\newblock Very deep convolutional networks for large-scale image recognition.
\newblock In {\em ICLR}, 2015.

\bibitem{srinivas2021bottleneck}
Aravind Srinivas, Tsung-Yi Lin, Niki Parmar, Jonathon Shlens, Pieter Abbeel,
  and Ashish Vaswani.
\newblock Bottleneck transformers for visual recognition.
\newblock In {\em CVPR}, 2021.

\bibitem{szegedy2015going}
Christian Szegedy, Wei Liu, Yangqing Jia, Pierre Sermanet, Scott Reed, Dragomir
  Anguelov, Dumitru Erhan, Vincent Vanhoucke, and Andrew Rabinovich.
\newblock Going deeper with convolutions.
\newblock In {\em CVPR}, 2015.

\bibitem{tolstikhin2021mlp}
Ilya Tolstikhin, Neil Houlsby, Alexander Kolesnikov, Lucas Beyer, Xiaohua Zhai,
  Thomas Unterthiner, Jessica Yung, Andreas Steiner, Daniel Keysers, Jakob
  Uszkoreit, et~al.
\newblock Mlp-mixer: An all-mlp architecture for vision.
\newblock {\em arXiv preprint arXiv:2105.01601}, 2021.

\bibitem{touvron2021training}
Hugo Touvron, Matthieu Cord, Matthijs Douze, Francisco Massa, Alexandre
  Sablayrolles, and Herv{\'e} J{\'e}gou.
\newblock Training data-efficient image transformers \& distillation through
  attention.
\newblock In {\em ICML}, 2021.

\bibitem{touvron2021going}
Hugo Touvron, Matthieu Cord, Alexandre Sablayrolles, Gabriel Synnaeve, and
  Herv{\'e} J{\'e}gou.
\newblock Going deeper with image transformers.
\newblock {\em arXiv preprint arXiv:2103.17239}, 2021.

\bibitem{tran2015learning}
Du Tran, Lubomir Bourdev, Rob Fergus, Lorenzo Torresani, and Manohar Paluri.
\newblock Learning spatiotemporal features with 3d convolutional networks.
\newblock In {\em ICCV}, 2015.

\bibitem{tran2019video}
Du Tran, Heng Wang, Lorenzo Torresani, and Matt Feiszli.
\newblock Video classification with channel-separated convolutional networks.
\newblock In {\em ICCV}, 2019.

\bibitem{tran2018closer}
Du Tran, Heng Wang, Lorenzo Torresani, Jamie Ray, Yann LeCun, and Manohar
  Paluri.
\newblock A closer look at spatiotemporal convolutions for action recognition.
\newblock In {\em CVPR}, 2018.

\bibitem{vaswani2017attention}
Ashish Vaswani, Noam Shazeer, Niki Parmar, Jakob Uszkoreit, Llion Jones,
  Aidan~N Gomez, {\L}ukasz Kaiser, and Illia Polosukhin.
\newblock Attention is all you need.
\newblock In {\em NIPS}, 2017.

\bibitem{wang2020video}
Heng Wang, Du Tran, Lorenzo Torresani, and Matt Feiszli.
\newblock Video modeling with correlation networks.
\newblock In {\em CVPR}, 2020.

\bibitem{wang2021tdn}
Limin Wang, Zhan Tong, Bin Ji, and Gangshan Wu.
\newblock Tdn: Temporal difference networks for efficient action recognition.
\newblock In {\em CVPR}, 2021.

\bibitem{wang2021pyramid}
Wenhai Wang, Enze Xie, Xiang Li, Deng-Ping Fan, Kaitao Song, Ding Liang, Tong
  Lu, Ping Luo, and Ling Shao.
\newblock Pyramid vision transformer: A versatile backbone for dense prediction
  without convolutions.
\newblock {\em arXiv preprint arXiv:2102.12122}, 2021.

\bibitem{wang2018non}
Xiaolong Wang, Ross Girshick, Abhinav Gupta, and Kaiming He.
\newblock Non-local neural networks.
\newblock In {\em CVPR}, 2018.

\bibitem{wang2021action}
Zhengwei Wang, Qi She, and Aljosa Smolic.
\newblock Action-net: Multipath excitation for action recognition.
\newblock In {\em CVPR}, 2021.

\bibitem{wu2021cvt}
Haiping Wu, Bin Xiao, Noel Codella, Mengchen Liu, Xiyang Dai, Lu Yuan, and Lei
  Zhang.
\newblock Cvt: Introducing convolutions to vision transformers.
\newblock {\em arXiv preprint arXiv:2103.15808}, 2021.

\bibitem{xie2017aggregated}
Saining Xie, Ross Girshick, Piotr Doll{\'a}r, Zhuowen Tu, and Kaiming He.
\newblock Aggregated residual transformations for deep neural networks.
\newblock In {\em CVPR}, 2017.

\bibitem{xie2018rethinking}
Saining Xie, Chen Sun, Jonathan Huang, Zhuowen Tu, and Kevin Murphy.
\newblock Rethinking spatiotemporal feature learning: Speed-accuracy trade-offs
  in video classification.
\newblock In {\em ECCV}, 2018.

\bibitem{yao2021seco}
Ting Yao, Yiheng Zhang, Zhaofan Qiu, Yingwei Pan, and Tao Mei.
\newblock Seco: Exploring sequence supervision for unsupervised representation
  learning.
\newblock In {\em AAAI}, 2021.

\bibitem{yu2021s}
Tan Yu, Xu Li, Yunfeng Cai, Mingming Sun, and Ping Li.
\newblock S$^{2}$-mlpv2: Improved spatial-shift mlp architecture for vision.
\newblock {\em arXiv preprint arXiv:2108.01072}, 2021.

\bibitem{zhang2021token}
Hao Zhang, Yanbin Hao, and Chong-Wah Ngo.
\newblock Token shift transformer for video classification.
\newblock In {\em ACM MM}, 2021.

\bibitem{zhang2018shufflenet}
Xiangyu Zhang, Xinyu Zhou, Mengxiao Lin, and Jian Sun.
\newblock Shufflenet: An extremely efficient convolutional neural network for
  mobile devices.
\newblock In {\em CVPR}, 2018.

\bibitem{zhang2021vidtr}
Yanyi Zhang, Xinyu Li, Chunhui Liu, Bing Shuai, Yi Zhu, Biagio Brattoli, Hao
  Chen, Ivan Marsic, and Joseph Tighe.
\newblock Vidtr: Video transformer without convolutions.
\newblock In {\em ICCV}, 2021.

\bibitem{zhong2020random}
Zhun Zhong, Liang Zheng, Guoliang Kang, Shaozi Li, and Yi Yang.
\newblock Random erasing data augmentation.
\newblock In {\em AAAI}, 2020.

\bibitem{zoph2017neural}
Barret Zoph and Quoc~V Le.
\newblock Neural architecture search with reinforcement learning.
\newblock In {\em ICML}, 2017.

\end{thebibliography}
}

\end{document}